\pdfoutput=1
\documentclass[conference]{IEEEtran}
\IEEEoverridecommandlockouts

\usepackage{pbalance}

\IEEEoverridecommandlockouts 
\usepackage{graphicx}
\usepackage{algorithmic}
\usepackage{algorithm}
\usepackage[T1]{fontenc}
\usepackage[cmex10]{amsmath}
\usepackage{amsfonts}
\usepackage{amssymb}
\usepackage{url}
\usepackage{multirow}
\usepackage{tablefootnote}
\usepackage[numbers]{natbib}

\title{Improving Domain-Specific Retrieval by NLI Fine-Tuning}

\author{
\IEEEauthorblockN{Roman Dušek}
\IEEEauthorblockA{Allegro sp. z o.o. \\
Wierzbięcice 1B, 61-569 Poznań, Poland \\
Email: roman.a.dusek@allegro.com}
\and
\IEEEauthorblockN{Aleksander Wawer}
\IEEEauthorblockA{0000-0002-7081-9797\\
* Allegro sp. z o.o. \\ 
Wierzbięcice 1B, 61-569 Poznań, Poland \\
** Institue of Compter Science, Polish Academy of Sciences \\
Jana Kazimierza 5, 01-248 Warszawa \\
Email: ** axw@ipipan.waw.pl, * aleksander.wawer@allegro.com}
\and
\IEEEauthorblockN{Christopher Galias, Lidia Wojciechowska}
\IEEEauthorblockA{Allegro sp. z o.o. \\
Wierzbięcice 1B, 61-569 Poznań, Poland \\
Email: \{krzysztof.galias,lidia.wojciechowska\}@allegro.com}
}

\begin{document}
\maketitle              

\begin{abstract}
The aim of this article is to investigate the fine-tuning potential of natural language inference (NLI) data to improve information retrieval and ranking. We demonstrate this for both English and Polish languages, using data from one of the largest Polish e-commerce sites and selected open-domain datasets. We employ both monolingual and multilingual sentence encoders fine-tuned by a supervised method utilizing contrastive loss and NLI data. Our results point to the fact that NLI fine-tuning increases the performance of the models in both tasks and both languages, with the potential to improve mono- and multilingual models. Finally, we investigate uniformity and alignment of the embeddings to explain the effect of NLI-based fine-tuning for an out-of-domain use-case.

\end{abstract}

\section{Introduction}

Query and sentence embedding vectors are used in information retrieval to match the searched query to results, for example in ranking of the results returned by lexical search engines \cite{detext} or in vector-based similarity search \cite{johnson2019billion}. 

The standard approach to training text encoders is to use large-scale corpora such as Wikipedia or CommonCrawl and the Masked Language Modeling (MLM) objective. A setup like this was used to train HerBERT \cite{mroczkowski-etal-2021-herbert}, the state-of-the-art monolingual BERT for the Polish language, which utilized Polish-specific datasets and the Sentence Structural Objective in addition to MLM. CommonCrawl, Wikipedia, and MLM were also used to train XLM-RoBERTa \cite{conneau-etal-2020-unsupervised}, a transformer supporting 100 languages.

In past years there have been numerous applications of natural language inference (NLI) data in training large language models such as sentence encoders. One example supporting the Polish language is the multilingual Universal Sentence Encoder (USE) \cite{yang2019multilingual}. For the 16 covered languages, training data included question-answer pairs, translation pairs, and the SNLI~\cite{snli} corpus, translated using Google Translate into target languages. The model was trained in a dual encoder setup and comes in two variants: a lightweight convolutional neural network and a transformer.

Recently, NLI data were applied in a combination with contrastive loss in a method called SimCSE \cite{gao-etal-2021-simcse}. It demonstrated superior performance on STS (Semantic Textual Similarity) tasks. Contrastive fine-tuning was also reported to improve ranking quality when applied to multilingual encoders \cite{litschko_cross-lingual_2022}. 

Unfortunately, large NLI datasets suitable for model training are usually not available in languages other than English. For this reason, in this work we test the feasibility of using machine translated NLI data and demonstrate this approach for Polish. We will use both monolingual (Polish and English) and multilingual models and evaluate them on data in both languages.

In this paper, we focus on two information retrieval tasks: the retrieval task, which aims to find a set of documents that match the query, and the ranking task, which sorts the results by relevance to the query. To demonstrate the proposed approach, our experiments will be performed on out-of-domain models, by which we mean generic, pre-trained neural language models that have not been tuned to real-world search data such as user clicks. We explore the impact of using translated NLI data for contrastive fine-tuning. We consider how does the fine-tuning affect information retrieval and ranking tasks. Furthermore, we investigate whether the uniformity and alignment of embeddings are linked to out-of-domain information retrieval performance.

The paper is organized as follows: in Section~\ref{sec:experiments} we introduce datasets and experimental setup, Section~\ref{sec:results} discloses results and Section~\ref{sec:discussion} concludes the paper by drawing conclusions.

\section{Experiments}\label{sec:experiments}
\subsection{Datasets}

We examined the performance of the models on three types of benchmarks. 

The first one is not directly related to information retrieval. This is a generic approach to evaluate pre-trained large neural language models. The first part is based on a GLUE-like collection for testing the selected model on a number of downstream benchmarks. We use it in Polish, where such a benchmark is the KLEJ framework \cite{klej2020}. In our paper we report averaged model performance on KLEJ datasets. The second part consists of semantic textual similarity (STS) tasks:
\begin{itemize}
    \item translated SICK-R \cite{dadas_evaluation_2020} available from the Polish version of SentEval\footnote{\url{https://github.com/sdadas/polish-sentence-evaluation}},
    \item CDS-R \cite{wroblewska-krasnowska-kieras-2017-polish}, a Polish dataset based on SICK-R,
    \item translated STSB\footnote{\url{https://huggingface.co/datasets/stsb\_multi\_mt/viewer/pl/train}}.
\end{itemize}
These datasets contain pairs of sentences human labelled based on the relatedness. 

The second benchmark is ranking using a random sample consisting of 86K search listings from one of the largest e-commerce platforms in Poland. The listings consist of a search phrase and the first page of results (on average 50 offers) from the lexical search engine along with information about the clicked items. We sorted the listings according to the cosine similarity between the embedding of the search phrase and the embedding of each offer title. We assessed the performance of the models by calculating click-based NDCG and averaging the results.

The third benchmark consists of two retrieval tasks. Here we applied Polish monolingual and multilingual models used in previous benchmarks, but also English monolingual models to extend our research to other languages. To evaluate Polish models in the retrieval task, we used an internal dataset from one of the largest e-commerce Polish platforms, which consists of search results. It is a sample of 30K user queries and 1M product titles, containing at least one clicked product for each of the user queries. English language models were tested on two datasets. The first one is WANDS \cite{wands}, a similar dataset from the e-commerce domain. Its test subset contains 379 queries and 43K candidate products with human-labelled query-product pairs. The main purpose is evaluation of semantic search in e-commerce. To broaden our evaluation, we further tested English models on the second English dataset, outside of e-commerce, namely SciFact \cite{scifact}. It is included in BEIR \cite{beir}, an information retrieval benchmark. SciFact's test subset contains 300 scientific claims (queries) verified against a corpus of 5K abstracts. 

\subsection{NLI translation}
We evaluated the translations using COMET (Crosslingual Optimized Metric for Evaluation of Translation) \cite{rei-etal-2020-comet} scores, an automated method of assessing translation quality. COMET is a new neural framework for evaluating multilingual machine translation models. COMET is designed to predict human judgments of machine translation quality. We used the older model, namely wmt20-comet-qe-da\footnote{\url{https://github.com/Unbabel/COMET}} to compare the translation results. The newer COMET release has a better correlation with human evaluation and a less skewed distribution of scores, but the calculated values were more difficult to interpret and establish a threshold value that indicates good vs bad translation quality.

The mBart\footnote{\url{https://huggingface.co/facebook/mbart-large-50-one-to-many-mmt}} model reached score a of $0.49$ compared to $0.40$ of m2m100\footnote{\url{https://huggingface.co/docs/transformers/model_doc/m2m_100}}, which is why we decided to translate the data using mBart. We also experimented with choosing the best of two translations for each sentence, which we comment on later in Section~\ref{sec:Rq3}.

\subsection{Training details}
We selected several models for fine-tuning with the supervised SimCSE framework\footnote{\url{https://github.com/princeton-nlp/SimCSE}}. In the case of Polish, we applied SimCSE to the Polish monolingual model HerBERT~\cite{mroczkowski-etal-2021-herbert}, which achieved top scores in the Polish KLEJ benchmark. In the case of English, we selected the English-only monolingual base variant of BERT (BERT-base-uncased) \cite{DevlinBERT}. Finally, we applied SimCSE to the multilingual model XLM-RoBERTa~\cite{conneau-etal-2020-unsupervised}, which also is the best multilingual model on the KLEJ leaderboard. We fine-tuned HerBERT and XLM-RoBERTa models using the SNLI dataset translated to Polish\footnote{We also tested a combination with MNLI, but this resulted in worse performance in information retrieval tasks.}, and the English SNLI and MNLI data in the case of English BERT and XLM-RoBERTa (in the case of English fine-tuning).

\subsection{SimCSE: Contrastive loss using NLI} 
SimCSE~\cite{gao-etal-2021-simcse} is a contrastive learning method aimed at generating sentence embeddings. First, it utilizes an unsupervised approach, which takes an input sentence and predicts itself in contrastive objective, with dropout used as noise. Authors find that dropout acts as minimal data augmentation, and removing it leads to a representation collapse. Then, they propose a supervised approach, which incorporates annotated pairs from natural language inference (NLI) datasets into the contrastive learning framework by using "entailment" pairs as positives and "contradiction" pairs as hard negatives. The contrastive loss is formulated for paired examples $D = \left \{ \left ( {x}_{i}, {x}^{+}_{i}   \right ) \right \} ^{m}_{i=1}$, where ${x}_{i}$ and ${x}^{+}_{i}$ are semantically related. Assuming that ${h}_{i}$ and ${h}^{+}_{i}$ are representations of ${x}_{i}$ and ${x}^{+}_{i}$, the training objective is:

$$ \ell_{contrastive} = -\log\frac{{\rm e}^{\text{sim}\left(\mathbf{h}_{i}, \mathbf{h}^{+}_{i}\right)/\tau}}{\sum^{N}_{j=1}{\rm e}^{\text{sim}\left(\mathbf{h}_{i}, \mathbf{h}^{+}_{j}\right)/\tau}}$$

where $\tau$ is a temperature hyperparameter and sim(${h}_{i}$, ${h}^{+}_{i}$) is the cosine similarity.

Following the SimCSE~\cite{gao-etal-2021-simcse} we used their supervised training framework to fine-tune selected models on SNLI dataset translated into Polish. This supervised task takes advantage of human-labelled pairs of sentences. As in the original work, we treated entailment pairs as positives and contradiction pairs as a hard negatives.

\subsection{Uniformity and alignment}
Wang et al.~\cite{wang2020} identify two key properties of embeddings, \emph{uniformity} and \emph{alignment}, and propose to use them to measure embedding quality. Later work~\cite{wang_towards_2022} in the recommender domain also suggests that better uniformity and alignment increases NDCG. Alignment is meant to measure whether similar samples have similar embeddings and is given by

$\ell_{align} \triangleq\mathbb{E}_{(x,y)\sim p_{pos}}\lVert f(x)-f(y) \rVert^\alpha_2,\quad\alpha>0,$

where $f$ is a function mapping an entity to its embedding and $p_{pos}$ is a distribution of positive pairs. Uniformity measures whether maximal information is preserved between the input and embedding space, which leads to spreading out of the representations, and is given by $$\ell_{uniform}\triangleq\log \mathbb{E}_{x,y\sim p_{data}}\left[e^{-t\lVert f(x) - f(y) \rVert^2_2}\right],\quad t>0,$$ where $p_{data}$ is the input distribution.

\section{Results}\label{sec:results}

\subsection{Results of SimCSE with translated NLI}
As we can see in Table~\ref{tab:results-sts-klej}, the role of SimCSE is ambiguous: it greatly improves the STS performance, but in the case of the best Polish monolingual model Herbert, it degrades its performance on the KLEJ benchmark. 

The results regarding STS and general benchmarks such as KLEJ agree with the observations of SimCSE authors in~\cite{gao-etal-2021-simcse}. They are somewhat selective: the focus is on evaluating SimCSE on semantic textual similarity (STS), and indeed in this benchmark their method performs in a competitive manner. However, the performance on many other typical downstream tasks, such as for example GLUE benchmark's sentiment analysis, is not competitive and is mentioned only in the appendix of the SimCSE paper. Authors conclude that sentence-level objective of SimCSE may not directly benefit such transfer tasks.

\begin{table*}[h]
\centering
\caption{Results of the STS and KLEJ evaluation tasks and number of supported languages (\#langs).}
\label{tab:results-sts-klej}
\begin{tabular}{lccccccc}
\hline
                   & STSB-PL & SICK-R & CDS-R & Avg STSB-PL & Avg KLEJ & \#langs                       \\ \hline
HerBERT            & 0.302   & 0.369  & 0.605 & 0.425   & \bf{86.3} & 1 \\
SimCSE-HerBERT     & 0.742   & \bf{0.781}  & 0.905 & \bf{0.809}   & 84.5 & 1   \\
XLM-RoBERTa        & 0.584   & 0.561  & 0.821 & 0.655   & 81.5 & 100\\
SimCSE-XLM-RoBERTa & 0.727   & 0.766  & 0.888 & 0.793   & 81.7 & 100\\
USE\tablefootnote{We used the transformer variant available at \\  \url{https://tfhub.dev/google/universal-sentence-encoder-multilingual-large/3}} & \bf{0.749} & 0.691  & \bf{0.909} &  0.783 & -\tablefootnote{KLEJ value cannot be computed for USE in a manner directly comparable to other solutions, because it supports only one input and does not support the `[SEP]` special tokens as the other transformer models do. Some of the KLEJ subsets are paired, as for example question-answer or paraphrase data.}   & 16 \\ \hline
\end{tabular}
\end{table*}

\subsection{Results of information retrieval benchmark}
 Table~\ref{tab:english retrieval} presents the results of the English benchmark. To get the best possible performance from used models we use both mean-pooling (average representation of tokens in sequence) and the CLS token representations. This doesn't discriminate against models which are not fine-tuned for utilisation of the CLS token (e.g. BERT). Tables~\ref{tab:polish ranking} and~\ref{tab:polish retrieval} show results of the Polish language tasks.
Generally, SimCSE fine-tuning improves both NDCG and recall. For both languages the best results in terms of retrieval, as reflected in Recall@100 scores, were obtained by monolingual BERTs with SimCSE fine-tuning. Except for the case of the English WANDS benchmark, USE was second in terms of performance, ahead of XLM-RoBERTa fine-tuned by SimCSE. In the ranking task HerBERT, SimCSE-HerBERT, and USE shared first place when using the mean of the last hidden layer to represent the utterance. In the CLS+pooler representation, SimCSE-HerBERT was the best one.

\begin{table*}[h]
\centering
\caption{Results of evaluation on retrieval task using English datasets. Numbers reported represent Recall@100.}
\begin{tabular}{ccccc}
\hline
                                               & \multicolumn{2}{c}{WANDS} & \multicolumn{2}{c}{BEIR-SciFact} \\ \hline
\multicolumn{1}{c}{Model / Inference Pooling} & mean            & CLS+pooler          & mean          & CLS+pooler       \\ \hline
\multicolumn{1}{c}{BERT-base-uncased}         & 0.2543          & 0.0632              & 0.5134        & 0.0200           \\
\multicolumn{1}{c}{SimCSE-BERT-base-uncased}           & 0.4933          & \bf{0.4991}              & \bf{0.7832}        & 0.6306           \\
\multicolumn{1}{c}{XLM-RoBERTa}                     & 0.1648          & 0.1458              & 0.1506        & 0.2368           \\
\multicolumn{1}{c}{SimCSE-XLM-RoBERTa}          & 0.3986          & 0.4338              & 0.5701        & 0.6878         \\
\multicolumn{1}{c}{USE}          & 0.3964          & -             & 0.7665        & -         \\ \hline
\end{tabular}

\label{tab:english retrieval}
\end{table*}

\begin{table}[]
\centering
\caption{Results of evaluation on ranking task in Polish. Numbers reported represent NDCG\tablefootnote{We computed statistical significance of averaged NDCGs using the paired T-test, p-value$<0.05$. Non-significant pairs where we could not confirm the differences were USE vs SimCSE-HerBERT and XLM-RoBERTa vs HerBERT. In other cases the differences are statistically significant.}.}
\begin{tabular}{ccc}
\hline
                                   & \multicolumn{2}{c}{Ranking test set} \\ \hline
\multicolumn{1}{c}{Model / Pooling}         & mean               & CLS+pooler            \\ \hline
\multicolumn{1}{c}{HerBERT}       & \bf{0.312}             & 0.307               \\
\multicolumn{1}{c}{SimCSE-HerBERT} & \bf{0.312}            & \bf{0.312}                \\ 
\multicolumn{1}{c}{XLM-RoBERTa}       & 0.306             & 0.305              \\
\multicolumn{1}{c}{SimCSE-XLM-RoBERTa}       & 0.309           & 0.309             \\
\multicolumn{1}{c}{USE}       & \bf{0.312}           & -             \\ \hline
\end{tabular}
\label{tab:polish ranking}
\end{table}

\begin{table}[]
\centering
\caption{Results of evaluation on retrieval task in Polish. Numbers reported represent Recall@100.}
\begin{tabular}{ccc}
\hline
                                   & \multicolumn{2}{c}{Retrieval test set} \\ \hline
\multicolumn{1}{c}{Model / Pooling}         & mean               & CLS+pooler            \\ \hline
\multicolumn{1}{c}{HerBERT}       & 0.0230             & 0.0222                \\
\multicolumn{1}{c}{SimCSE-HerBERT} & 0.2476             & \bf{0.2562}                \\ 
\multicolumn{1}{c}{XLM-RoBERTa}       & 0.0020             & 7.48e-5               \\
\multicolumn{1}{c}{SimCSE-XLM-RoBERTa}       & 0.1487             & 0.1621              \\
\multicolumn{1}{c}{USE}       & 0.2407            & -             \\ \hline
\end{tabular}

\label{tab:polish retrieval}
\end{table}


\subsection{Uniformity and alignment}
\begin{figure}[ht]
\centering
\includegraphics[width=0.9\linewidth]{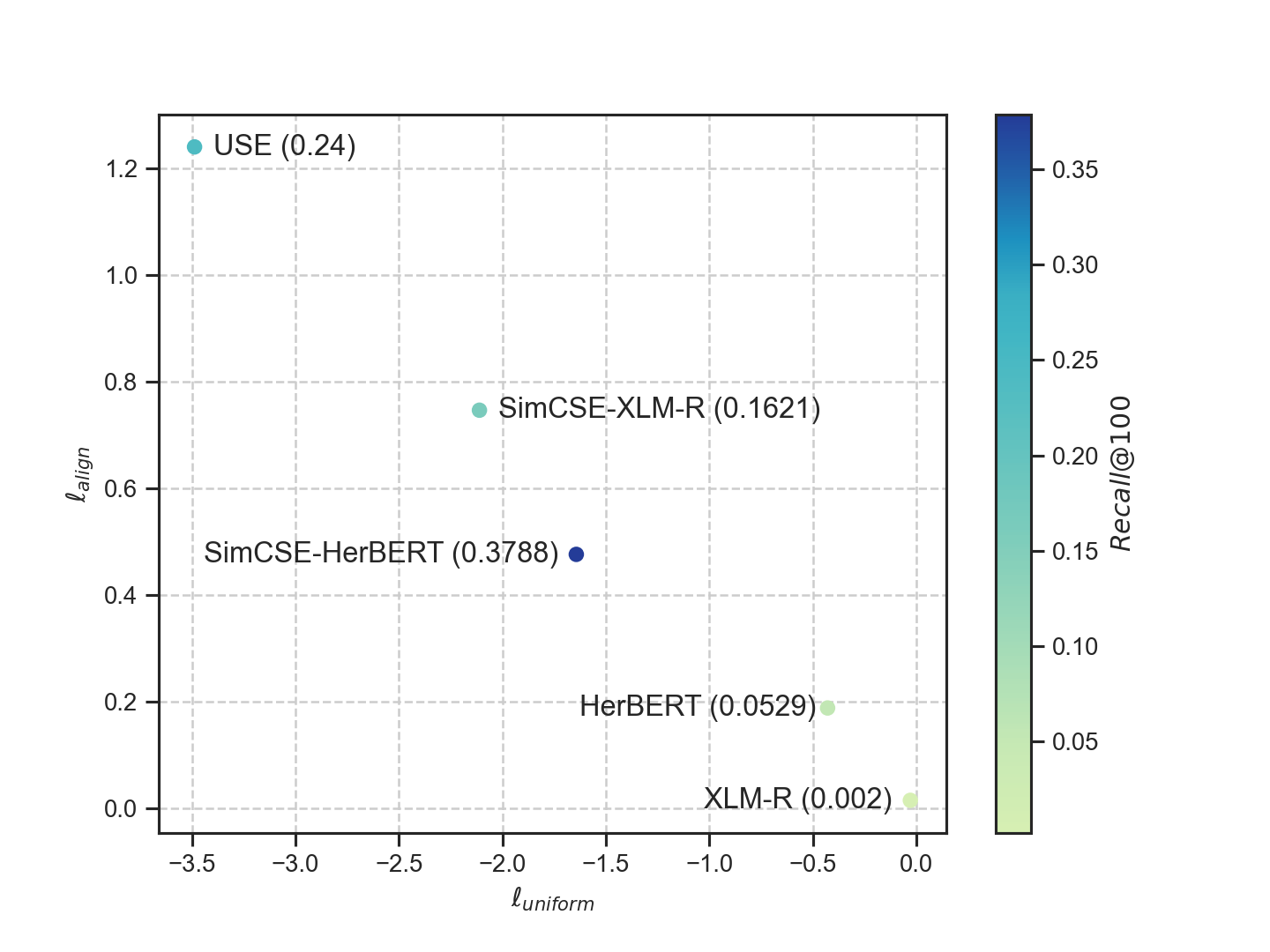}
\caption{Recall@100 on the plot of $\ell_{align}$ versus $\ell_{uniform}$ on vector-search dataset. For both axes lower is better. Colors and numbers in parentheses indicate Recall@100.}
\label{fig:align_unif_retrieval}
\end{figure}

 We calculated uniformity and alignment using the search phrase and title with a click, utilizing a batch size of $1024$ over 300K of pairs, with the default $\alpha=2$ and $t=2$. Contrastive fine-tuning improved the performance of both HerBERT and XLM-RoBERTa. However, only uniformity improved as the alignment metric increased (see Figure~\ref{fig:align_unif_retrieval}).

\subsection{Influence of translation quality}
\label{sec:Rq3}
In order to examine the influence of poorly translated sentences we conducted experiments where we filtered translated sentences based on the COMET score. Using both translation from mBart and m2m100 models, we selected the highest COMET score translation to pick one example from each of the translated datasets. The average COMET score on SNLI rose by $7$ percentage points after filtering. After inspecting the cleaned datasets many examples with scores close to zero were still found. Removing examples with scores lower than $0.05$ resulted in reducing the dataset size by $1/3$. Fine-tuning the model on the cleaned dataset resulted in worse performance than baseline.

\section{Discussion}\label{sec:discussion}
\label{sec:discuss}

Using the translated SNLI dataset had a comparable effect to the results reported in \cite{gao-etal-2021-simcse}. This confirms the role of translated NLI for improving the model performance, even despite possible translation errors.

The USE model competes with monolingual models when it comes to STS benchmarks. Contrastive loss, as applied in SimCSE, is not used in the USE model. Moreover, the USE model is multilingual, as it supports $16$ languages, and it contains only $80$ mln parameters in the large variant, compared to $110$ mln of the HerBERT and XLM-RoBERTa base versions. The only element that is common to both the USE and HerBERT with SimCSE fine-tuning is the usage of NLI data for model training. Therefore, we conclude that it is the NLI fine-tuning that plays the key role in information retrieval and STS performance.

Another interesting observation is that the averaged KLEJ score is not related to information retrieval capability. However, better performance on the semantic textual similarity tasks (STSB-PL, SICK-R and CDS-R) is. Our results demonstrate that SimCSE fine-tuning degrades monolingual model performance on the KLEJ benchmark, therefore it should not be considered as a one-size-fits-all method for tuning language models. We believe that using NLI data for model pre-training and/or fine-tuning has a positive effect in representing text for information retrieval problems.  

We observed a link between information retrieval and uniformity dimension only. We did not observe a relationship between alignment and information retrieval as is reported in~\cite{gao-etal-2021-simcse} or in the context of recommender systems~\cite{wang_towards_2022}. Previous work assessed alignment and uniformity using an in-domain setting, compared to our case of an out-of-domain scenario — but the impact of this setting concerning alignment remains an open research question.

All multilingual models scored higher on uniformity compared to monolingual models. We believe this is because multilinguality makes the model use more of the embedding space. Moreover, the alignment of all multilingual models was worse compared to monolingual models. This shows that alignment and uniformity do not directly translate to capabilities of sentence encoders.

\section{Conclusions and future work}
Our results show that state-of-the-art performance in out-of-domain retrieval and ranking tasks can be achieved with a method based on contrastive loss and NLI data, such as SimCSE, applied to a pre-trained language model. We confirm the positive effect of contrastive loss using both monolingual and multilingual models, pointing to the conclusion that the key to superior performance in out-of-domain information retrieval is fine-tuning sentence encoders using NLI data.

In this paper we did not train the model on clicks. This could be done using contrastive loss. In the future we plan to optimize sentence encoders on click data using alignment and uniformity in the loss function, as in~\cite{wang_towards_2022}.


%
%
%
\bibliographystyle{IEEEtran}
\bibliography{main}
\end{document}